%% file: acl2024.tex
\pdfoutput=1
\documentclass[11pt]{article}

\usepackage[]{ACL2024}

\usepackage{times}
\usepackage{latexsym}

\usepackage[T1]{fontenc}
\usepackage[utf8]{inputenc}

\usepackage{microtype}

\usepackage{inconsolata}
\usepackage{amsmath}
\usepackage{graphicx}
\usepackage{multirow}
\usepackage{latexsym}
\usepackage{amssymb}
\usepackage{makecell}
\usepackage{booktabs}
\usepackage{bbm}
\usepackage{wrapfig}
\usepackage{lipsum} % 仅用于生成示例文本
\usepackage{xcolor}
\usepackage{caption}

\usepackage{graphicx,calc}
\usepackage{wrapfig}
\newlength\myheight
\newlength\mydepth
\settototalheight\myheight{Xygp}
\title{Editing Conceptual Knowledge for Large Language Models}

\author{Xiaohan Wang$^\spadesuit$$^\diamondsuit$, Shengyu Mao$^\spadesuit$$^\diamondsuit$, Shumin Deng$^\clubsuit$, 
Yunzhi Yao$^\spadesuit$$^\diamondsuit$, Yue Shen$^\heartsuit$,\\ \textbf{Lei Liang$^\heartsuit$, Jinjie Gu$^\heartsuit$, Huajun Chen$^\spadesuit$$^\diamondsuit$, Ningyu Zhang$^\spadesuit$$^\diamondsuit$\thanks{~~Corresponding author.}}\\
$^\spadesuit$Zhejiang University,~ 
$^\diamondsuit$ZJU-Ant Group Joint Research Center for Knowledge Graphs,\\
$^\clubsuit$National University of Singapore, NUS-NCS Joint Lab, Singapore,~ $^\heartsuit$Ant Group\\
\texttt{\{wangxh07,zhangningyu\}@zju.edu.cn}\\
\raisebox{-\mydepth}{\includegraphics[height=1.6\myheight]{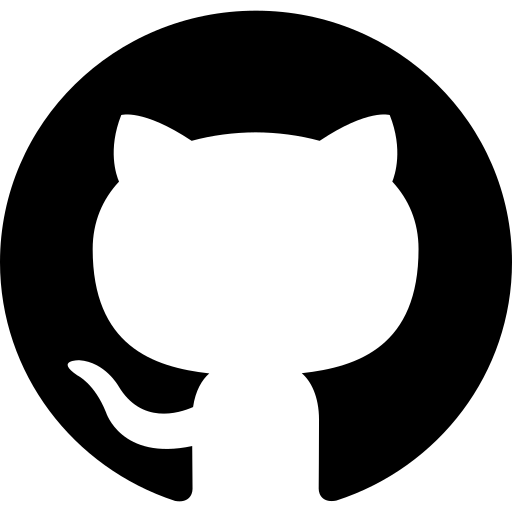}}
\textbf{\url{https://zjunlp.github.io/project/ConceptEdit}}
}

\begin{document}
\maketitle

\def\ours{ConceptEdit}
\def\metricone{Concept Consistency}
\def\metrictwo{Instance Change}
\begin{abstract}
Recently, there has been a growing interest in knowledge editing for Large Language Models (LLMs). Current approaches and evaluations merely explore instance-level editing, while whether LLMs possess the capability to modify concepts remains unclear. This paper pioneers the investigation of editing conceptual knowledge for LLMs, by constructing a novel benchmark dataset \textbf{\ours} and establishing a suite of new metrics for evaluation. The experimental results reveal that, although existing editing methods can efficiently modify concept-level definitions to some extent, they also have the potential to distort the related instantial knowledge in LLMs, leading to poor performance. We anticipate this work can inspire further progress in understanding LLMs.
% \footnote{Code is in \url{https://github.com/zjunlp/EasyEdit}.}
\end{abstract}

\section{Introduction}

\input{sections/introduction}

\section{Background}
\input{sections/background}

\section{Concept Editing}
\input{sections/concept_editing}
\section{Benchmark Construction}
\input{sections/dataset}
\input{sections/tab/main_table}

\section{Experiment}
\input{sections/experiment}

\section{Related Work}
\input{sections/related_work}
\section{Conclusion}
We introduce the conceptual knowledge editing task for LLMs, with a new benchmark \ours\ and evaluation metrics.
From the experiments, we observe that existing editing methods, when modifying conceptual knowledge, have a very limited impact on the underlying instances; thus, stronger techniques and better understandings of concepts in LLMs are necessary for further research.

\section*{Limitations}
Despite our best efforts, there remain several aspects that are not covered in this paper.
\paragraph{Models} Due to computation resource constrains, we could not incorporate larger-scale models or experiment with a wider variety of architectures, such as Vicuna~\citep{vicuna2023}, Qwen-72B~\citep{qwen}, Mixtral-8$\times$7B~\citep{DBLP:journals/corr/abs-2401-04088}.
These models garner interest within the community and remain to be explored in the future study.
\paragraph{Task Settings} About the scope of concept categorization presented herein, this paper delves into the realm of concrete concepts. 
However, it does not extensively cover the domain of abstract concepts, which encompass intangible entities or principles, such as rules and emotions \cite{DBLP:journals/corr/abs-2311-09174, DBLP:journals/corr/abs-2310-01405}. 
Editing these broader concepts, with their intrinsic complexity and subtlety, is beyond the confines of the current discussion and remains further research.
\paragraph{Mechanism}
This paper primarily analyzes the concept location and editing mechanisms within LLMs.   
The investigation into how LLMs learn and represent various concepts and entities, as well as the establishment of concept hierarchies, remains cursory. 
These aspects are yet to be fully understood and warrant more comprehensive study.

\section*{Ethics Statement}

This study adheres strictly to the most rigorous ethical standards and best practices in research.
All data utilized are extracted from datasets that are available to the public, thereby ensuring no usage of any proprietary or sensitive information. As a result, this research is free from any ethical concerns.

\section*{Acknowledgements}

We would like to express gratitude to the anonymous reviewers for their kind comments. 
This work was supported by the National Natural Science Foundation of China (No. 62206246, No. NSFCU23B2055, No. NSFCU19B2027), the Fundamental Research Funds for the Central Universities (226-2023-00138), Zhejiang Provincial Natural Science Foundation of China (No. LGG22F030011), Information Technology Center and State Key Lab of CAD\&CG, Zhejiang University, and NUS-NCS Joint Laboratory (A-0008542-00-00).

% \section*{Acknowledgements}
% We would like to express our sincere gratitude to DBpedia and Wikidata for providing open-source data that laid the groundwork for our research. 
% We are grateful to \citet{Wu_2023} for providing a dataset that served as a reference.
% Our appreciation also goes to ~\citet{DBLP:conf/nips/MengBAB22} for the casual tracing code utilized in our paper. 
% Their contributions are invaluable to the advancement of our work.

\bibliography{anthology,custom}
\bibliographystyle{acl_natbib}

\appendix
\clearpage
\section{Appendix}
\label{sec:appendix}
\subsection{Templates}
Here we numerates a variety of templates employed within our experimental framework.
% In recognition of the inherent abilities of LLMs to tackle reasoning tasks, we meticulously crafted the prompts to effectively leverage this capacity.

\begin{table}[ht]
\centering
\resizebox{1.0 \columnwidth}{!}{
\begin{tabular}{p{1.17\linewidth}}
\hline
\textbf{Template for Instance Change}  \\
\hline
% \textit{few-shot prompt:}\\
% \\
Whether FrancoAngeli belongs to category publisher? Yes\\
Whether And Other Stories belongs to category people? No\\
Whether [INSTANCE] belongs to category [CONCEPT]?\\
\hline
\end{tabular}
}
\caption{Template for \metrictwo}
\label{temp_ic}
\end{table}

Table \ref{temp_ic} shows the few-shot prompt used both before and after the edits. 
As introduced in Section \ref{sec:metric}, a positive response \texttt{yes} equates to a score of 1 in \metrictwo.
Upon revising the definitions of pertinent concepts, a shift in the instance-to-category relationship is anticipated. 
Thus, a negative response \texttt{no} from the model post-editing signifies that the relation is altered. 
For instance, when $I_{\theta}(C\in t)=1$ with $C$ representing concept `publisher' and initial definition, and $t$ referring to `Victoria University Press'.
Ideally, $I_{\theta_e}(t \in C^*)=0$, because the conceptual knowledge is changed, the post-edited model no longer associates `Victoria University Press' with the concept `publisher'.

\begin{table}[ht]
\centering
\resizebox{1.0 \columnwidth}{!}{
\begin{tabular}{p{1.2\linewidth}}
\hline
\textbf{Template for \metricone}  \\
\hline
Prediction sentence: [PREDICTION]\\
\\
Sentence A: [TARGET].\\
Sentence B: [GROUND].\\
\\
Check the prediction sentence and Give a score from -1 to 1:\\
Score 1: close meaning to sentence A\\
Score 0: neither relevant to A nor B\\
Score -1: close meaning to sentence B\\
\\
Output format is \{Score:\{\}, Reason:\{\}\}\\
\hline
\end{tabular}
}
\caption{Template for \metricone}
\label{temp_cg}
\end{table}
Table \ref{temp_cg} delineates the structured template utilized for the \textit{\metricone}, which acts as an input for the GPT-4 evaluator.
Through qualitative analysis, the \textit{\metricone} classifies the generated sentences into three discrete scores. 
The adoption of a relative comparison, rather than assigning an absolute value, acknowledges the proficiency of the evaluator that is preliminarily verified to align more closely with human judgment. 
% By instructing the response format, we utilize the reasoning steps to make the score more convincing and explainable, also easier to sparse.

\begin{table}[ht]
\centering
\resizebox{1.0 \columnwidth}{!}{
\begin{tabular}{p{1.12\linewidth}}
\hline
\textbf{Twenty Restructured Exemplars}  \\
\hline
The meaning of [CONCEPT] can be described as\\
In essence, [CONCEPT] is defined as\\
To put it simply, [CONCEPT] refers to

[CONCEPT] is characterized by the following definition\\
A concise explanation of [CONCEPT] is\\
Defined as such, [CONCEPT] can be understood as\\
When we talk about [CONCEPT], we mean\\
In simple terms, [CONCEPT] is defined as\\
The definition ascribed to [CONCEPT] is\\
To clarify, [CONCEPT] is defined by

[CONCEPT] is essentially defined as\\
Describing [CONCEPT], we can say\\
The definition assigned to [CONCEPT] is\\
In the context of [CONCEPT], we define it as\\
Putting it in words, [CONCEPT] is defined as\\
When we refer to [CONCEPT], we are talking about\\
In defining [CONCEPT], we consider it as\\
The characterization of [CONCEPT] involves\\
Defining [CONCEPT] boils down to\\
It can be stated that [CONCEPT] is defined as\\
\hline
\end{tabular}
}
\caption{Templates for equivalent neighbors}
\label{temp_eqnei}
\end{table}
Table \ref{temp_eqnei} enumerates twenty restructured exemplars derived from GPT-4 responses as the equivalent neighbors.
Equivalent neighbor means another sentence expressing similar semantic meaning of descriptor x, used in metric Generalization. 
Employing GPT-4 to reformulate ``\textit{The definition of [CONCEPT] is}'' facilitates the generation of equivalent neighbors varied and increased fluency.

\begin{table}[ht]
\centering
\resizebox{1.0 \columnwidth}{!}{
\begin{tabular}{p{1.2\linewidth}}
\hline
\textbf{Template for Method PROMPT}  \\
\hline
Prompt:\\
Definition of [CONCEPT]: [DESCRIPTOR Y]\\
\\
Example:\\
\textbf{Pre-Editing:}
The definition of military person is \textit{someone who rides horses in horse racing or steeplechase racing.}\\

\textbf{Post-Editing:}
Definition of military person: someone who rides horses in horse racing or steeplechase racing.\\
The definition of military person is \textit{someone who rides horses in horse racing or steeplechase racing.}\\
\hline
\end{tabular}
}
\caption{Template for Method PROMPT}
\label{temp_prompt}
\end{table}

The PROMPT method utilizes a prefix sentence as the prompt used for inference in LLM to instruct (edit) the output.
Table \ref{temp_prompt} presents the template employed by the PROMPT method for editing, along with an illustrative example showing the difference between pre-editing and post-editing sentences in a practical application.
This \textit{Prompt} portion is what constitutes the PROMPT method.
Furthermore, in the computation of metric \textit{Instance Change}, the \textit{Prompt} prefix is positioned antecedent to the few-shot demonstrations.
\subsection{Data Distribution}
\label{appendix:datadistribution}

\begin{table}[ht]
\centering
\resizebox{1.0 \columnwidth}{!}{
\begin{tabular}{p{0.8\linewidth} c}
\hline
Property& Number\\
\hline
        Number of \textit{Concepts} & 452\\
        Number of \textit{Instances} & 8,767\\
        Number of \textit{Superclasses} &  22\\
        Average tokens length of \textit{Description}&  12.95\\
        Max/Min tokens length of \textit{Description}&  45 / 3\\
\hline
\end{tabular}
}
\caption{\ours\ Dataset Statistics}
\label{Appendix:datastatis}
\end{table}
Table \ref{Appendix:datastatis} introduces the statistics of \ours\ dataset that describes its composition. 

\begin{figure*}
    \centering
    \includegraphics[width=1.0\linewidth]{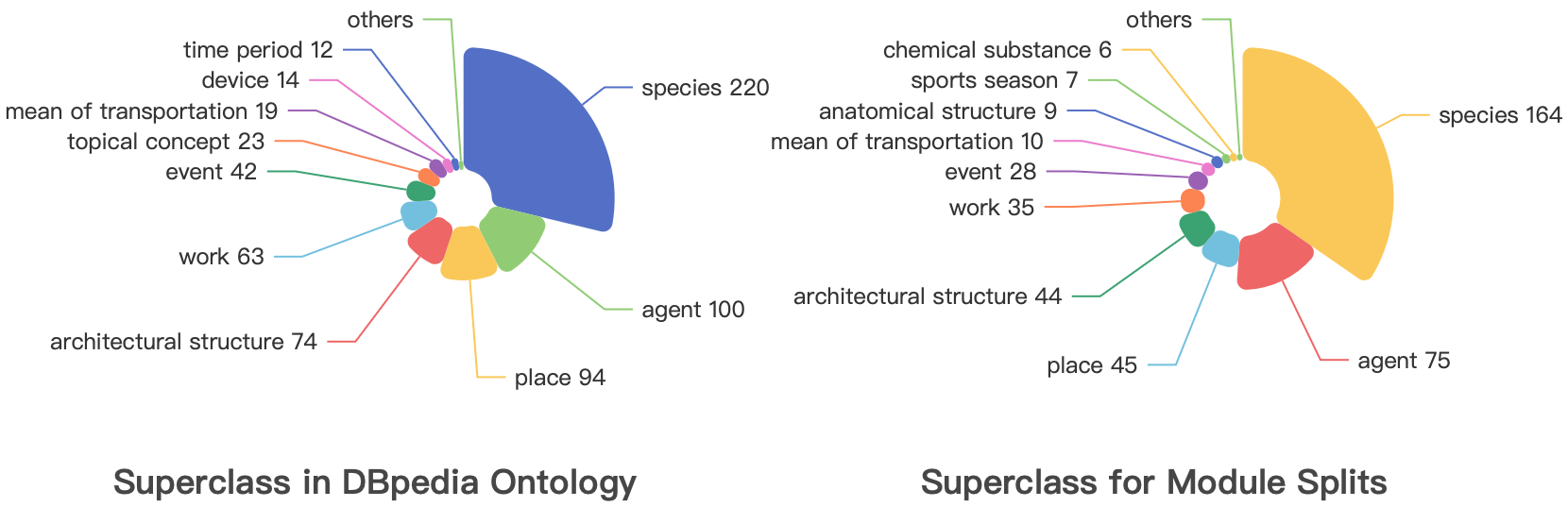}
    \caption{Statistics of superclass distribution. Considering DBpedia Ontology exhibits a hierarchical and tree-like arrangement, we category each concept based on its highest-level node. The left panel illustrates the frequency distribution among the original \textbf{DBpedia Ontology} concepts, whereas the right panel depicts the distribution of selected concepts in \textbf{\ours}.}
    \label{fig:superclasses}
\end{figure*}

Figure \ref{fig:superclasses} illustrates that the predominant superclass distribution of \ours\ bears a resemblance to origin ontology.
When dividing between intra and inter modules, we randomly pick a replacement concept either from the same group or a different one. 
For categories with fewer than five concepts, we make selection from the entire set.

\begin{figure*}
    \centering
    \includegraphics[width=1.0\linewidth]{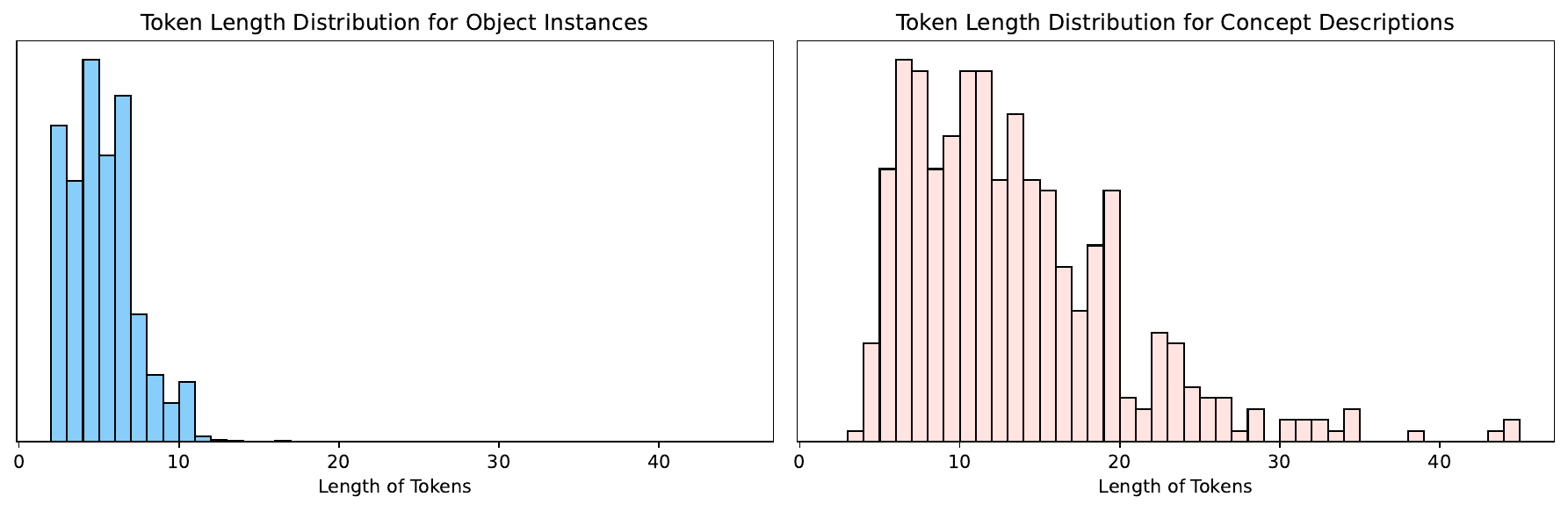}
    \caption{Comparison of tokens length tokenized by \llama\ for editing tasks.  
    The left table presents token lengths for the \textbf{zsRE} dataset, showing that most lengths fall below 10. On the right side, the table illustrates that the \textbf{\ours} dataset features tokens of greater length, encompassing a broader range of length. zsRE is commonly employed in instance-level editing, focusing on specific entities, while \ours\ involves editing descriptions for concepts, which tend to be more extensive.
    }
    \label{fig:token_length}
\end{figure*}

Figure \ref{fig:token_length} presents a comparison of the length of tokens between the prior editing dataset zsRE and our dataset. 
At the same time, Table \ref{Appendix:datasetvscounter} supplements our dataset and CounterFact dataset in terms of content differences. 
This is to illustrate the distinction between the conceptual knowledge editing task and the instance-level factual editing.

\input{sections/tab/cases}

Drawing on ontology datasets, our study utilizes the knowledge from DBpedia as performed by \citet{Wu_2023}. 
This initial phase involves the careful retrieval of a total of 783 distinct classes, each representing a specific concept within the ontology. 
The dataset not only retains the hierarchy of superclasses but also harnesses SPARQL to interrogate 20 instances chosen at random via the \texttt{type\_of} relation.
OntoProbe provides a solid foundation with elements such as concept names and their instances. 
However, to integrate it into proposed concept editing task, we recognize that substantial effort is required to adapt the dataset accordingly. 
Table \ref{Appendix:datasetvswu} demonstrates that our task with the OntoProbe dataset is entirely different.

\begin{table*}[]
\centering
\resizebox{2.08\columnwidth}{!}{
\begin{tabular}{|c|p{9.8cm}|c|p{3.2cm}|}
\hline
 &  \multicolumn{1}{c|}{\textbf{ConceptEdit}}&  &  \multicolumn{1}{c|}{\textbf{ CounterFact}}  \\ \hline
id & 1 & case\_id & 8 \\ \hline
\multirow{2}{*}{concept\_name} & \multirow{2}{*}{military person} & \multirow{2}{*}{prompt} & What is the twin city of Wellington? It is \\ \hline
concept\_def & those who serve as part of an organized armed military force & target\_new & Sheffield \\ \hline
top\_superclass & species & ground\_truth & Sydney \\ \hline
\multirow{8}{*}{instances} & {[}"Ronald Reid-Daly", "Charles Augustus Hilton","27th Indiana Infantry Regiment","Spartaco   Schergat", "Charles Corcoran", "Franois Claude Amour,   marquis de Bouill", "Clyde A. Vaughn", "Earle   Wheeler", "Joe McCarthy", "Central African Republic Civil   War", "Andrew Mathews", "Nikolaus Heilmann",   "Ahmed Abdel Rahman Nasser", "Reed McKinley Chambers",   "Wallace Lawler", "Clarence Tan", "Louis Charles   mile Gibon-Guilhem", "Manshuk Mametova", "Moshe   Tzadok", "John F. G. Howe"{]} & \multirow{8}{*}{rephrase\_prompt} & People in Wellington's twin city speak the language of \\ \hline
\multirow{2}{*}{QID} & \multirow{2}{*}{Q47064} & \multirow{2}{*}{locality\_prompt} & What is the twin city of Chicago? It is \\ \hline
\multirow{2}{*}{module\_intra}& \{"replace\_from\_concept": "jockey", "replace\_def":     "someone who rides horses in horse racing or steeplechase racing"     \} & \multirow{2}{*}{locality\_ground\_truth} & \multirow{2}{*}{Sydney} \\ \hline
\multirow{2}{*}{module\_inter} & \{"replace\_from\_concept":"settlement","replace\_def":   "place of any size, in which people live"\} &  &  \\ \hline
locality\_prompt & The definition of bacteria is &  &  \\ \hline
locality\_answer & domain of micro-organisms &  &  \\ \hline
\end{tabular}
}
\caption{The existing knowledge editing datasets differ significantly from ours; the current factual editing datasets are instance-level and lack exploration at the level of conceptual knowledge. We identify this gap during our preliminary research and thus transform an ontology dataset to create \ours. We list the differences between \ours\ and the commonly used factual editing dataset, CounterFact.}
\label{Appendix:datasetvscounter}
\end{table*}

\begin{table*}[]
\centering
\resizebox{2.08\columnwidth}{!}{
\begin{tabular}{|c|p{9.3cm}|c|p{7.8cm}|}
\hline
 &  \multicolumn{1}{c|}{\textbf{ConceptEdit}}&  &  \multicolumn{1}{c|}{\textbf{OntoProbe}}  \\ \hline
id & 1 & id & 1 \\ \hline
concept\_name & military person & rdfs:label & military person \\ \hline
concept\_def & those who serve as part of an organized armed military force & rdfs:Class & http://dbpedia.org/ontology/MilitaryPerson \\ \hline
top\_superclass & species & rdfs:subClassOf & ["person","animal","eukaryote","species"] \\ \hline
\multirow{11}{*}{instances} & {[}"Ronald Reid-Daly", "Charles Augustus Hilton", "27th Indiana Infantry Regiment", "Spartaco   Schergat", "Charles Corcoran", "Franois Claude Amour,   marquis de Bouill", "Clyde A. Vaughn", "Earle   Wheeler", "Joe McCarthy", "Central African Republic Civil   War", "Andrew Mathews", "Nikolaus Heilmann",   "Ahmed Abdel Rahman Nasser", "Reed McKinley Chambers",   "Wallace Lawler", "Clarence Tan", "Louis Charles   mile Gibon-Guilhem", "Manshuk Mametova", "Moshe Tzadok", "John F. G. Howe"{]} & \multirow{11}{*}{is rdf:type of} & ["Ronald Reid-Daly", "Charles Augustus Hilton", "27th Indiana Infantry Regiment", "Spartaco Schergat", "Charles Corcoran", "Franois Claude Amour, marquis de Bouill", "Clyde A. Vaughn", "Earle Wheeler", "Joe McCarthy (RCAF officer)", "Central African Republic Civil War (2012-present)", "Andrew Mathews", "Nikolaus Heilmann", "Ahmed Abdel Rahman Nasser", "Reed McKinley Chambers", "Wallace Lawler", "Clarence Tan", "Louis Charles mile Gibon-Guilhem", "Manshuk Mametova", "Moshe Tzadok", "John F. G. Howe"] \\  \hline
QID & Q47064 &  & \\ \hline
\multirow{2}{*}{module\_intra}& \{"replace\_from\_concept": "jockey", "replace\_def":     "someone who rides horses in horse racing or steeplechase racing"     \} &  &  \\ \hline
\multirow{2}{*}{module\_inter} & \{"replace\_from\_concept":"settlement","replace\_def":   "place of any size, in which people live"\} &  &  \\ \hline
locality\_prompt & The definition of bacteria is &  &  \\ \hline
locality\_answer & domain of micro-organisms &  &  \\ \hline
\end{tabular}
}
\caption{Here we showcase a comparation of \ours\ and OntoProbe example. Our dataset introduces concepts and applies them to editing tasks, whereas OntoProbe focuses more on exploring the structure of ontological knowledge. For example, our task needs to redefine those concepts, which entails the gathering of definition contexts from WIKIDATA. Furthermore, to employ calculation of editing metrics, we construct equivalent neighbors and out-of-scope neighbors needed. Those efforts are detailed in the paper section \ref{section:dataprocess}.}
\label{Appendix:datasetvswu}
\end{table*}

\subsection{Experiment Details}
\label{appendix:experiment}
We utilize four editing baselines on concept editing task, which are detailed as following:

\textbf{Finetune (FT)}
updates parameters by gradient descent for a single MLP layer and applies early stop strategy to constrain the modifications in the weights. Here, we adopt FT-M in EasyEdit \cite{DBLP:journals/corr/abs-2308-07269} which finetune a single layer by cross-entropy loss optimization.

\textbf{ROME} \citep{DBLP:conf/nips/MengBAB22} envisages the MLP module as a key-value storage, leverages causal mediation analysis to locate the edit area, and update a whole FFN layer to encode new knowledge.

\textbf{MEMIT} \citep{meng2023massediting} adopts the localization techniques in ROME and uses explicitly computed parameter updates to embed new memories across multi-layers.

\textbf{PROMPT} is well known that a well-designed prompt can effectively guide the behavior of LLMs, demonstrating a strong ability to learn from context. The prompt used here is shown in Table \ref{temp_prompt}.

The experimental procedures undertaken in this study are underpinned by the utilization of the tool \textbf{EasyEdit}\footnote{\url{https://github.com/zjunlp/EasyEdit}} \citep{DBLP:journals/corr/abs-2308-07269}.
Moreover, the selection of hyper-parameters adheres to the default configurations as established.
Taking into account the scale of \textit{LLaMA-2-7B-Chat} and \textit{Mistral-7B-v0.1}, we conduct our experiments on an A800 GPU within a local computing environment, for current editing methods involving more than just inference.

It is imperative to note that in concept editing task, the editing manipulation is performed dependently, targeting only the specified descriptor $(x_e, y_e)$ with a single edit at a time not sequentially. 
After the evaluation is completed for each sample, the edited model is reset to its original state before the edit. 
This ensures that each editing operation is isolated and does not affect subsequent edits, allowing for a controlled assessment of each individual modification to the conceptual knowledge.

\subsubsection{Special Circumstances of Instance Change on \gptj}
\label{appendix:special}
As introduced in \S~\ref{paragraph:instance}, we delegate a mandate to LLMs to generate an alternative instance that is expected to be within the confines of the model's existing corpus of knowledge to ensure continuity.
This step exists uncertainty, as there is a possibility that the newly generated instance might not satisfactorily address the query in Table \ref{temp_ic} before editing.
Despite our efforts to refine the demonstrations and retain an appropriate instance, we cannot assure that the function $G_{\theta}$ will always yield score 1.
The negative numbers in Table~\ref{tab:maintable} show shortage of GPT-J when recognizes such instance-to-category relationship.
It is somewhat peculiar to discover certain instances that initially fail to pass the identification process, yet post-editing, they oddly begin to affirm the query \texttt{yes}, which result in the negative number recorded in method PROMPT.
Anyhow, such situation is not commonly found in the other LLMs' performance in our experiment.

\subsubsection{Manual and GPT-4 Evaluation on Metric \metricone}
\label{Appendix:manual}

\begin{table}[ht]
\centering
\resizebox{0.95 \columnwidth}{!}{
\begin{tabular}{@{}lcc@{}}
\toprule
\metricone\ & GPT-4 & Human \\ \midrule
\textbf{-1}: close to origin & 19 & 16 \\
\textbf{0}: neither & 6 & 9 \\
\textbf{1}: close to target & 25 & 25 \\ \midrule
total number & 50 & 50\\
\bottomrule
\end{tabular}
}
\caption{Comparison of GPT-4 Scores vs. Human Scores on metric \metricone.}
\label{Appendix:ccgh}
\end{table}

To illustrate the effectiveness of using the GPT-4 API as an automatic evaluator for metric \metricone\, and to verify the extent to which GPT-4 evaluation of semantic similarity aligns with human evaluation, we sample 10 examples each from the pre-edited output and the outputs edited by four different editing methods, resulting in a total of 50 cases for comparison between GPT-4 and human evaluation results. The scoring criteria for the human evaluation are the same as those designed for the GPT-4 evaluation. Also, a graduate student assists us with the human evaluation.
The results, presented in Table \ref{Appendix:ccgh}, indicate that out of all 50 samples, only 3 records differ between GPT-4 and the human judges, with errors mainly occurring in the judgment of scores -1 and 0. 
However, for scores of 1 (that is, close to the target), all judgments are consistent. 
This implies that using GPT-4 as an automatic evaluator is reliable for metric \metricone.

\begin{figure*}
    \centering
    \includegraphics[width=1\linewidth]{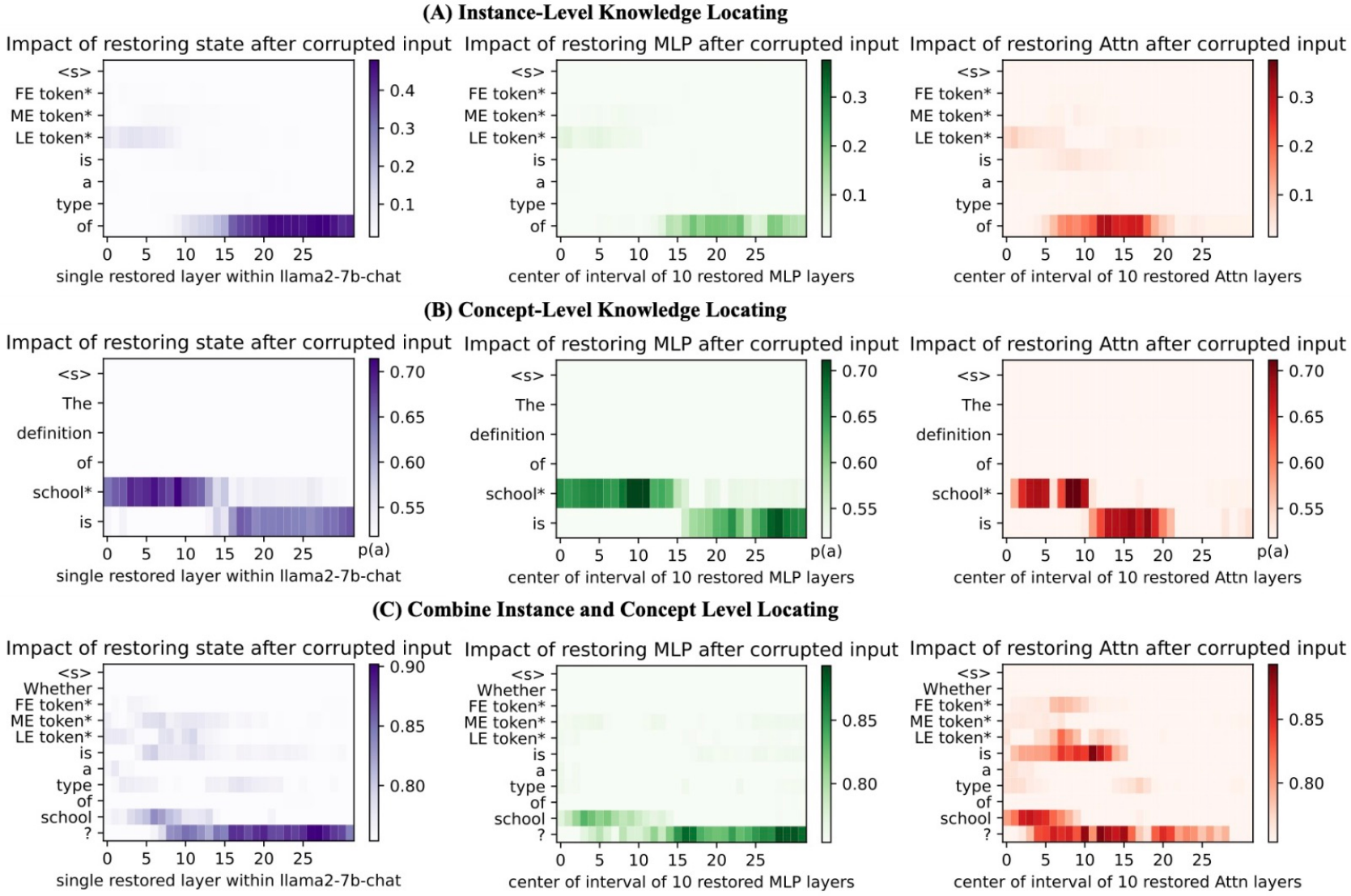}
    \caption{Case of casual tracing on concept ``\textit{school}'' shows similar appearance of the lookup patterns. As our previous study discussed, the conclusion is basically in line with case \textit{publisher}.}
    \label{fig:enter-label}
\end{figure*}

\subsection{Case in Generated Sentence}
\label{appendix:cases}

Table \ref{tab:cases} exhibits five representative cases of the generated sentences, showcasing the varying levels of success and failure in carrying out the edits.

\textit{CASE A: Ideal Successful Edit} In the best scenarios, the edited Sentence A perfectly aligns with the target sentence, with every word matching without any discrepancies. These are the desired outcomes.

\textit{CASE B: Consistent Meaning but Not a Perfect Match}  In some cases, the edited sentence, while not identical to the target sentence, conveys a similar core meaning. This could involve the use of synonyms or synonymous expressions. In human reviews, such cases are considered acceptable because they retain the main content.

\textit{CASE C: Partially Consistent but Differing in Meaning} There are also cases, where the edited sentence partially overlaps with the target sentence, but does not convey the exact same meaning, possibly differing in the explanation of certain key information. Although the result is not completely accurate, it is closer to the target than before editing, thus providing a point that is worth more attention in future research.

\textit{CASE D: Edit Failures But Original Meaning Maintained} In cases of editing failure, this is a typical example. In these situations, although the editing task was not successful, the model-generated sentences maintained their original semantic content without any substantive change.

\textit{CASE E: Neither Target Nor Original Meaning }Finally, we also discover special cases, a kind of editing failure where the generated sentence neither matches the editing target nor retains the original meaning. This situation is different from Case B because it does not have any consistency with the target nor does it maintain the meaning before editing, presenting an entirely unexpected result that also warrants further analysis and study.

\subsection{Knowledge Locating}\label{Appendix:knowledge_locating}

We introduce the knowledge locating details which is introduced by \citet{DBLP:conf/nips/MengBAB22}.
Given a model $f_{\theta}$ and an input text $X = \{x_i|i\in[1,N]\}$, where $N$ is the number of input tokens, and we denote the tokens to be perturbed as $x_t$, which refer to the subject entity.

\paragraph{Clean run} involves a normal forward process $f_{\theta}(X)$, and then saves the hidden activations $\{h_{i}^{l}|i\in[1,N],l\in[1,L]\}$, here $L$ indicates the layer number of model $f_{\theta}$. 

\paragraph{Corrupted run.} Then we conduct the process of corrupting. Specifically, after embedding the tokens as $\{h_{i}^{0}|i\in[1,N]\}$,
% The hidden embedding of $x_t$ at different layer can be denoted as $[h_t^{1},h_t^{2},...,h_t^{L}]$. 
we directly add noise to the entity tokens $x_t$ before they are fed into the model, denoted as $h_t^{0}:=h_t^{0}+\epsilon$.
Here $\epsilon \sim N(0,\nu) $, and we follow previous work~\citep{DBLP:conf/nips/MengBAB22} to select $\nu$ to be 3 times larger than that of the empirical standard deviation of embeddings. In this way, we obtain the corrupted hidden activations $\{h_{i*}^{l}|i\in[1,N],l\in[1,L]\}$.

\paragraph{Corrupted-with-restoration run} hooks the model $f_\theta$ and iteratively attempts to restore the corrupted hidden state at each token and each layer to the clean state without intervening the future computations. 

\paragraph{Indirect Effect.} The final probability on target tokens of the three runs above is defined as $\mathbb{P}$, $\mathbb{P}_{*}$ and $\mathbb{P}_{*}^{clean~h_i^{l}}$. The indirect effect (IE) of a particular hidden state $h_i^{l}$ is defined as $\text{IE}=\mathbb{P}_{*}^{clean~h_i^{l}}-\mathbb{P}_{*}$.

\end{document}

%% file: sections/introduction.tex
The emergence of Large Language Models (LLMs) represents a significant step towards the era of AGI, with the performance of large-scale models being evident for all to see \citep{DBLP:journals/corr/abs-2303-12712,DBLP:journals/corr/abs-2303-18223}.
Despite their advancements, LLMs encounter challenges such as misinformation, outdated knowledge due to the training cut-off, and the risk of producing toxic content \cite{DBLP:journals/corr/abs-2310-05189, DBLP:journals/corr/abs-2310-07521, DBLP:journals/corr/abs-2307-12966, DBLP:conf/emnlp/ZhangFCNW23, DBLP:journals/corr/abs-2401-05561, feng2023trends, DBLP:journals/corr/abs-2310-19852}.
Since retraining LLMs to address these issues is time-consuming  and costly, there is a surge necessity for advancements in knowledge editing methods designed for LLMs, which facilitate efficient, post-training adjustments to the models \cite{DBLP:journals/corr/abs-2312-05497,DBLP:journals/corr/abs-2310-19704, DBLP:journals/corr/abs-2310-16218,zhang2024comprehensive,wei2024stable,DBLP:journals/corr/abs-2402-08631,DBLP:journals/corr/abs-2402-11324,peng2024event}. 
Besides, sparse autoencoders could generate interpretable features for LLMs' behavior~\cite{templeton2024scaling,gao2024scaling}.
Recent knowledge editing methods can achieve the instance-level editing ability to alter knowledge in LLMs.
Yet, such a case-by-case setting of knowledge editing is highly inefficient and lacks modeling of relations between instances.

\begin{figure}
    \centering
    \includegraphics[width=0.95\linewidth]{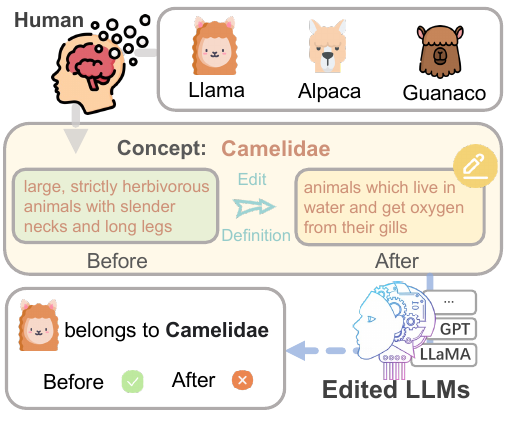}
    \caption{Humans learn conceptual knowledge from concrete instances and these concepts can guide further learning. Conceptual knowledge editing focuses on  modifying the definition of concepts to achieve conceptual knowledge modification in LLMs, and investigates the Top-Down Influence on instances.}
    \label{fig:conceptediting}
\end{figure}

Cognitive science \cite{HOLZINGER2023100788,zhao2023model,e2e750876f2d4663aa99df93fe35fe0f,rane2024concept} has revealed that humans understand new things and acquire new knowledge through learning concepts.
For example, the concept \textit{Camelidae} is \textit{large, strictly herbivorous animals with slender necks and long legs}.  
This abstraction, derived from concrete instances like \textit{llama} and \textit{alpaca}, assists in categorizing new entities.
Humans can achieve updates to a large amount of instances through concepts, thus, the \textit{llama} \textbf{DOES NOT} belong to \textit{Camelidae} anymore.
The distinctiveness of human leads to the research question: \textbf{whether LLMs learn and update concepts analogously}~\citep{lv2024coggpt,lo2024large,suresh2023conceptual}, as well as how to encapsulate and update concepts within parametric framework~\cite{DBLP:conf/acl/OnoeZPDC23,jamali2023unveiling}.

To this end, we propose \textbf{\ours}\footnote{CC BY-NC-SA 4.0 license.}, a novel benchmark dataset for editing conceptual knowledge, which tries to modify the definition of concepts in LLMs.
\ours\ is constructed upon the foundation of DBpedia Ontology \citep{10.1007/978-3-540-76298-0_52}, a widely recognized and cross-domain ontology that preserves conceptual knowledge hierarchically.
We build concepts with corresponding definitions and associated instances, accompanied by necessary elements for editing. 
Except for the common metrics for instance-level editing, we design two concept-specific metrics, \metrictwo\ for top-down influence on instances and \metricone\ for semantic similarity of generated definition.
Experiments with FT, ROME, MEMIT, and PROMPT methods show that recent knowledge editing baselines can reach high reliability in distorting concept-level definitions for LLMs, but still perform poorly on concept-specific metrics.

% We further analyze the conceptual knowledge mechanism, highlighting the challenges involved in editing concepts.
Moreover, conceptual operation represents a higher dimension of pre-training models, distinct from the learning through demonstrations of individual instances. 
Concept editing allows for efficient updates by generalizing from one to many or implementing controllable content generation through abstract expression interventions.  
It changes the model's understanding of abstract affairs through the manipulation of a concept, thereby achieving more efficient model control.

In conclusion, our investigation leads to a collection of interesting findings, where we highlight the following contributions:
\begin{itemize}
    \item We define a new task of conceptual knowledge editing for LLMs and construct a benchmark dataset, \textbf{\ours}.
    \item Furthermore, we develop a suite of metrics to evaluate the efficacy of current editing baselines on  conceptual knowledge editing. 
    New metrics, including \metrictwo\ and \metricone, are tailored to better show the capabilities of existing methods.
    \item By employing scenarios of concept distortion, we seek to unveil the underlying mechanisms how LLMs store and manage these concepts from the perspective of knowledge editing.
\end{itemize}

%% file: sections/background.tex
%In order to efficiently address the problems during the practice of LLMs, the methods of knowledge editing come into being \citep{feng2023trends}.
The objective of knowledge editing is to rectify particular factual inaccuracies encountered, without retraining the foundational model, while emphasizing the preservation of unrelated knowledge to the greatest extent, as elucidated by \citet{DBLP:conf/emnlp/CaoAT21}.
% In a formulaic definition of knowledge editing tasks, t
The given edit descriptor $(x,y)$ symbolizes the pairs that denote inputs and corresponding outputs embedded in LLMs. 
The base model $f_\theta$ undergoes an extensive learning to assimilate the edited $(x_e,y_e)$, ultimately producing an edited model as $f_{\theta_{e}}$.
% To achieve control, $x_e$ and $y_e$ need to be concatenated to maximize the conditional probability, as shown in Equation \ref{conditional probability}:
% \begin{equation}
% {\theta}_e = \operatorname{argmax}_{\theta} P(y_e | x_e; \theta)
% \label{conditional probability}
% \end{equation} 
To achieve this goal, $x_e$ and $y_e$ need to be concatenated to maximize the conditional probability, formally expressed as ${\theta}_e = \operatorname{argmax}_{\theta} P(y_e | x_e; \theta)$.

At present, a burgeoning interest in exploring the capabilities of knowledge editing exists~\citep{DBLP:journals/corr/abs-2309-08952, DBLP:journals/corr/abs-2401-17623, ma2024possible, DBLP:journals/corr/abs-2401-17809, chen2024can}, with the goal of developing more advanced methodologies.
% Most works are directed towards the rectification of factual inaccuracies, given that these represent the most common scenarios with which LLMs are confronted.
% These research efforts primarily concentrate on modifying factual knowledge within LLMs typically at the instance level.
These researches primarily concentrate on modifying factual knowledge typically at the instance level, encompassing various aspects.
Factual knowledge datasets, like zsRE~\citep{DBLP:conf/emnlp/CaoAT21} and CounterFact~\citep{DBLP:conf/nips/MengBAB22}, are frequently used as benchmarks~\citep{DBLP:journals/corr/abs-2401-07453,  li2023pmet, DBLP:journals/corr/abs-2312-11795}.
zsRE, context-free question-answering dataset, uses  rephrasings generated by backtranslation as the equivalence neighborhood and train/val splits.
For example, the answer of ``Which continent is Mount Andrewson?'' is changed to ``South America''.
CounterFact identifies between superficial changes in model word selections from specific and generalized alterations.
Employing triples as an external knowledge repository through their conversion into natural language is favored because relational datasets offer more definitive query responses, enhancing convenience in evaluations.
% These can be proficiently integrated to investigate the impact of such modifications on instance-level facts or commonsense knowledge. 
These can be proficiently integrated to see how changes affect instance-level facts.

%% file: sections/concept_editing.tex
\subsection{Task Definition}
Concept \cite{mckenna-etal-2021-multivalent,KDD2021_AliCG,DBLP:journals/dint/JiWSZWY19,DBLP:conf/aaai/GongZZ16,10.1145/2213836.2213891} is a generalization of the world, which represents the shared features and essential characteristics of a class of entities.
Concept editing aims to modify the definition of concepts, thereby altering the behavior of LLMs when processing these concepts.

In this study, the notation $C=(c, d)$ is employed to encapsulate a concept, where $c$ is the name of concept (e.g. publisher), and $d$ means the definition of concept (e.g. company that prints and distributes pressed goods or electronic media).
From the perspective of knowledge representation, concept editing for LLMs is concerned with the alteration of the extant $C = (c, d)$ into a modified representation $C^* = (c, d^*)$, in which $d^*$ corresponds to the revised definition.
In this manner, $c$ forms the basis of $x_e$, providing the necessary context for concept editing, and similarly, $d^*$ lays the foundation for $y_e$ in optimization.

Moreover, the notation $t$ denotes concrete instances (e.g. Victoria University Press) of the aforementioned concept. 
Here, $t \in C$ is employed to formally signify that the specific entity belongs to the broader category represented by the concept.
This membership relation, denoted by `$\in$', is frequently referred to as the `is\_a' relation or alternatively as the `is\_type\_of' relation. 
When editing conceptual knowledge, it is important to figure out the impact of this relationship and how it may be altered as a result of such conceptual changes.

\subsection{Metrics}
\label{sec:metric}
To analyze conceptual knowledge modification, we adopt the  metrics for factual editing (the target is the concept $C$ rather than factual instance $t$), adhering to the framework established by \citet{DBLP:conf/emnlp/YaoWT0LDC023}.
Although concept editing shares some commonalities with other factual editing tasks, our empirical investigations reveal that extant metrics fall short in offering a fine-grained assessment of changes to instance associations. 
Besides, given the length of definition text, a verbatim comparison of tokens emerges as an inadequate approach. 
Consequently, we devise novel metrics tailored to more accurate measurement for concept editing.

\textbf{\metrictwo.}
We present a detailed check of current editing techniques through instances.
Recognizing a gap in the precise quantification of instance-level changes, we develop an innovative metric capturing the intricacies of these alterations. 
This new metric \metrictwo\ is formulated as:
\begin{equation}
    \mathbb{I} (I_{\theta}(t \in C)-I_{\theta_e}(t \in C^*))
\end{equation}
where the function $I(t \in C^*)$ is defined such that it gets value 1 when the instance $t$ belongs to concept $C^*$ in the edited model and conversely, it adopts value 0 when $t \notin C^*$.
This categorization utilizes the reasoning ability of LLMs with prompt in Table \ref{temp_ic}, offering a nuanced understanding of their potential on instance-level modification.

\textbf{\metricone.}
This metric evaluates the semantic similarity of generated concept definition, which upon manual inspection correspond to three distinct scenarios, calculated as :
\begin{equation}
    H( g, d^*, d ) = \{1, 0, -1\}
\end{equation}
The generated text $g$ (concept definition) after edits delivers pertinent information that verifies the accurate editing of concepts.
In the scoring criterion, score 1 indicates high resemblance with the target definition; -1 denotes greater resemblance to the original definition; and 0 reflects ambiguity between them.
For automatic evaluation, we deploy GPT-4 API \citep{openai2023gpt4} as the evaluation model, which shows greater alignment with human preferences.
We also choose several cases for manual review in Appendix \ref{Appendix:manual}.
The evaluator as $H(.)$ generates responses based on prompts crafted according to a specific template in Table \ref{temp_cg}.

%% file: sections/dataset.tex
\begin{figure*}
    \centering
    \includegraphics[width=0.95\linewidth]{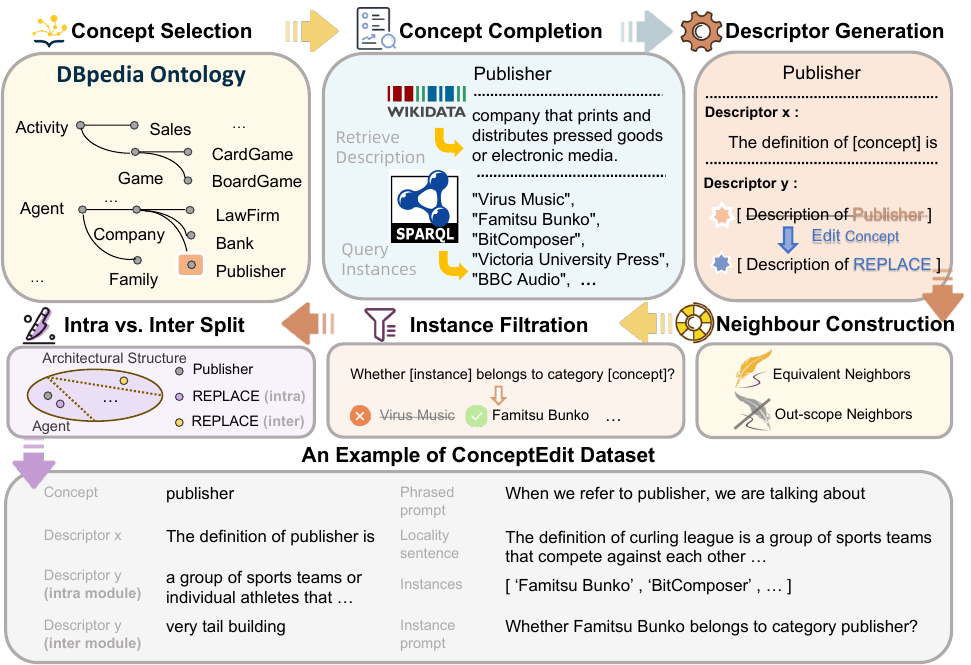}
    \caption{Overview of \textbf{\ours} benchmark construction. Building on the DBpedia Ontology, we enrich concepts with detailed definitions and associated instances, ensuring quality through meticulous processes.}
    \label{fig:dataset}
\end{figure*}
\subsection{Concept Selection}
It is widely acknowledged that ontology is a formal representation of concepts and represents highly structured knowledge~\citep{DBLP:conf/acl/0008CJ0023}. 
Classes, the focus of ontology, include a series of individual instances in a systematic manner.  
Therefore, our benchmark \ours\ incorporates the DBpedia ontology \citep{10.1007/978-3-540-76298-0_52}, a tree-like structure, to assemble a collection of concepts.

Drawing from the OntoProbe dataset by \citet{Wu_2023}, we collect concepts and their corresponding instances. 
However, our focus is on updating conceptual knowledge rather than constructing ontological structures as OntoProbe does.
Then, classes without instances are excluded to ensure integrity, resulting in a refined collection of 452 classes. 
Owing to the lack of definitions in DBpedia, we turn to Wikidata~\citep{DBLP:journals/cacm/VrandecicK14}, another well-regarded and freely available knowledge base, to augment our dataset with essential descriptive content.
To ensure data quality, \textbf{we manually review all the descriptions we gathered, replacing any unclear or ambiguous}.

\subsection{Data Processing}
\label{section:dataprocess}
\paragraph{Descriptor Generation.}

We initiate our descriptor generation process by a manually curated template to transform single concept name to natural language text for LLMs, serving as the $x$ component in our descriptor pair. 
The template adheres to a pre-defined formula: ``The definition of [Concept name] is''. 
Subsequently, we embark on using a distinct concept strategically chosen to supplant the original definition, thereby constructing the target $y$ component. 
For instance, ``very tall building'', which comes from the definition of concept ``skyscraper'', might be utilized as a substitute.

\paragraph{Neighbour Construction.}   

When descriptor undergoes editing, its equivalent neighbor, another sentence that expresses a similar idea, should also be edited accordingly. 
We construct twenty restructured sentences as inputs in metric Generalization to increase the flexibility of the equivalent neighbors, as demonstrated in Table \ref{temp_eqnei}.
Meanwhile, its out-scope neighbour for metric Locality is ascertained through a randomized selection mechanism from the pool of remaining, unaffiliated concepts.
% This checks the integrity of other concepts preserved, preventing unwarranted alterations.build equivalent neighbors for metric Generalization and out-scope neighbors for metric Locality

\paragraph{Instance Filtration.} 
\label{paragraph:instance}

Instances are carefully examined to ensure that LLMs possess relevant prior knowledge about the concepts and instances under investigation. 
This is executed through a binary evaluation mechanism denoted as $I_\theta(C\in  t)$, wherein a definitive determination is made based on the question: "Whether [instance] belongs to category [concept]?" 
To ensure that LLMs can make such judgments, we use the ``few-shot'' approach \citep{DBLP:conf/nips/BrownMRSKDNSSAA20}.
Additionally, if LLMs are unable to understand any instances retrieved from DBPedia, they are directed to create an alternative instance. 
This is a contingency strategy to address any gaps in knowledge that may arise from data repositories for different LLMs.

\paragraph{Intra vs. Inter split.}

We redefine concept A by employing the definition of concept B.
In this context, our data is divided into two splits.
One is Intra module: a prefix meaning ``within'' or ``inside'', that concept B is within the same superclass as concept A.
This implies concepts A and B share a higher-level relationship, which is expected to be easily aligned. 
Intra split assesses the effectiveness of concept editing in a relatively less challenging setting. 
In contrast, the Inter module selects concept B from the separate superclass, suggesting that the two concepts are less connected and their definitions are likely to be more divergent.
% By using these splits, the experiment can more thoroughly investigate the versatility and limits of concept editing. 

\subsection{Data Statistics}

Finally, we obtain \textbf{\ours}, containing 452 concepts, 8,767 instances with 22 superclasses.
The overview of the benchmark construction is shown in Figure \ref{fig:dataset}.
For detailed statistics and comparisons with prior datasets, see Appendix \ref{appendix:datadistribution}.

%% file: sections/tab/main_table.tex
\begin{table*}[ht]
\centering
% \LARGE
\resizebox{2.0\columnwidth}{!}{
\begin{tabular}{ll|cccc|cccc}

\toprule
\multirow{2}{*}{\textbf{Base Model}} & \multirow{2}{*}{\textbf{Method}} & \multicolumn{4}{c|}{\textbf{Intra}} & \multicolumn{4}{c}{\textbf{Inter}} \\
& & \textbf{Reliability$\uparrow$} & \textbf{Gen.$\uparrow$} & \textbf{Locality$\uparrow$} & \textbf{Inst.$\uparrow$} & \textbf{Reliability$\uparrow$} & \textbf{Gen.$\uparrow$} & \textbf{Locality$\uparrow$} & \textbf{Inst.$\uparrow$} \\
\midrule
% \multicolumn{10}{c}{\textsc{GPT-series}} \\
% \midrule
\multirow{4}{*}{\textit{GPT2-XL}}
&FT &    69.18    &  38.51      &  78.96  &\underline{17.70}    & 66.11& 35.30 & 77.72 & \underline{17.48}\\
&ROME &   \underline{86.47}     & \underline{49.68}     &  \underline{84.86}   &  \textbf{23.01}  & \underline{82.85}    & \underline{45.51}    &   \underline{86.21}  &  \textbf{20.13} \\
&MEMIT  &  51.07      &  35.48    &  \textbf{95.50}   & 3.32    & 46.35& 32.18 &\textbf{95.27} & 3.98 \\
&PROMPT  &   \textbf{88.26}     &  \textbf{86.30}      &   70.54  &  4.42  & \textbf{88.54}& \textbf{86.24}&  70.59 & 3.54\\
\midrule
\multirow{4}{*}{\textit{GPT-J-6B}}
&FT &   \textbf{100.0}     &  \textbf{92.76}       &  57.86   & \textbf{19.25}  & \textbf{100.0} & \textbf{92.56} & 59.05 & \textbf{22.34}\\
&ROME &  99.20    &  83.01    &   \underline{70.14}  & \underline{14.16}   & 99.21& 81.94& \underline{71.07} & 13.27\\
&MEMIT  &  \underline{99.83}      & 59.84  &   \textbf{94.20}  & 13.05  & \underline{99.55}& 56.15 & \textbf{94.80}& \underline{15.27}\\
&PROMPT   & 88.41 & \underline{86.42} & 69.10 &-18.14 & 88.66    &  \underline{87.01}  & 70.14   &   -17.70 \\
\midrule
\multirow{4}{*}{\parbox{2cm}{\textit{LLaMA-2-7B-Chat}}}
&FT  &  \textbf{100.0}      &  \textbf{89.60}   &  84.53   &  0.66 & \textbf{100.0} & \textbf{89.07} &85.49 & 0.44\\
&ROME    &   \underline{92.46}     &   70.92  &   \textbf{92.75}  &  \textbf{32.74}  & \underline{91.83}  &  71.16 &  \textbf{92.87}&  \underline{34.51}\\
&MEMIT  &   91.18   &   78.47   &  \underline{89.89}   &  \underline{30.75}   & 90.92  &  77.92  &  \underline{91.37}   &  \textbf{35.62}\\
&PROMPT  &  89.20      &  \underline{87.38}      & 76.92    &  3.76   & 88.74& \underline{87.89} & 77.77 & 2.21\\
\midrule
\multirow{4}{*}{\parbox{2cm}{\textit{Mistral-7B-v0.1}}}
&FT  &   \textbf{100.0}    &  76.16  & \textbf{95.83}   & 0.0 & \textbf{100.0}  & 72.98  & \textbf{96.31}& 0.0\\
&ROME   &   \underline{96.47}    &   76.11   &  \underline{93.99}   & \underline{10.62} & \underline{96.56} &  76.00&  \underline{94.37}& \underline{11.06}\\
&MEMIT  &  95.24     &  \underline{78.42}    &   91.97  &  \textbf{16.81} & 95.31 &\underline{76.98}  & 91.20   & \textbf{15.93}\\
&PROMPT  &   90.22    &   \textbf{88.65}   &  81.31   & 0.44 & 90.17 & \textbf{88.68} & 82.75 & 0.22\\
\bottomrule
\end{tabular}
}
\caption{Main results of the baselines on the \ours. \textbf{Bold} results denote the best performance in each setting, while \underline{underlined} results signify the second-best. $\uparrow$ means the metric goes higher if it performs better.
\textbf{Gen.} is the abbreviation of metric Generalization and \textbf{Inst.} is the abbreviation of metric \metrictwo.}
\label{tab:maintable}
\end{table*}

%% file: sections/experiment.tex
\def\llama{\textbf{LLaMA-2-7B-Chat}}
\def\gptxl{\textbf{GPT2-XL}}
\def\gptj{\textbf{GPT-J}}
\def\mistral{\textbf{Mistral-7B-v0.1}}

% model
\subsection{Experimental Setting}
\paragraph {Language models} 
Four most prevalent open-source LLMs are used as base models for editing tasks.
More precisely, we respectively utilize {\gptj} (6B) \citep{gpt-j}, {\gptxl} (1.5B) \citep{radford2019language}, {\llama} \citep{touvron2023llama} and {\mistral} \citep{jiang2023mistral} across various autoregressive models.

% method
\paragraph{Methods}
We select four distinct methodologies commonly used for knowledge editing, namely:
FT, ROME \citep{DBLP:conf/nips/MengBAB22}, MEMIT \citep{meng2023massediting} and PROMPT.
Detailed descriptions of these methods are presented in  Appendix \ref{appendix:experiment}.

% metrics
\paragraph{Evaluation Metrics.} 
To measure the impact of concept editing, we established a series of metrics,
some following the setup by \citet{DBLP:conf/emnlp/YaoWT0LDC023}:

\textbf{Reliability.} This metric straightforwardly measures the mean accuracy on a specific collection of pre-defined input-output pairs $(x_e,y_e)$:
\begin{equation}
\mathbb{E}_{x_{\mathrm{e}}^{\prime}, y_{\mathrm{e}}^{\prime} \sim \left\{\left(x_{\mathrm{e}}, y_{\mathrm{e}}\right)\right\}} \mathbbm{1} \left\{\operatorname{argmax}_y f_{\theta_{e}}\left(y \mid x_{\mathrm{e}}^{\prime}\right)=y_{\mathrm{e}}^{\prime}\right\}
\end{equation}

\textbf{Generalization.} Considering that paraphrased sentences should be modified accordingly by editing, this metric gauges the average accuracy on equivalent neighbors $R(x_e,y_e)$:
\begin{equation}
\mathbb{E}_{x_{\mathrm{e}}^{\prime}, y_{\mathrm{e}}^{\prime} \sim R\left(x_{\mathrm{e}}, y_{\mathrm{e}}\right)} \mathbbm {1} \left\{\operatorname{argmax}_yf_{\theta_{e}}\left(y \mid x_{\mathrm{e}}^{\prime}\right)=y_{\mathrm{e}}^{\prime}\right\}
\end{equation}

\textbf{Locality.}
Noted as specificity within some literature,  
this metric is assessed based on the frequency at which the predictions of the post-edit model remain unchanged in out-scope neighbors $O(x_e,y_e)$:
\begin{equation}
\mathbb{E}_{x_{\mathrm{e}}^{\prime}, y_{\mathrm{e}}^{\prime} \sim O\left(x_{\mathrm{e}}, y_{\mathrm{e}}\right)} \mathbbm {1} \left\{f_{\theta_{e}}\left(y \mid x_{\mathrm{e}}^{\prime}\right)=f_{\theta}\left(y \mid x_{\mathrm{e}}^{\prime}\right) \right\}
\end{equation}

\textbf{Concept Specific Evaluation Metrics}
We also utilize \textbf{\metrictwo} and  \textbf{\metricone}  introduced in \S \ref{sec:metric}, revealing the instance level variation and  semantic similarity of generated concept definition for concetual knowledge editing.

% results
\begin{figure*}
    \centering
    \includegraphics[width=0.98\linewidth]{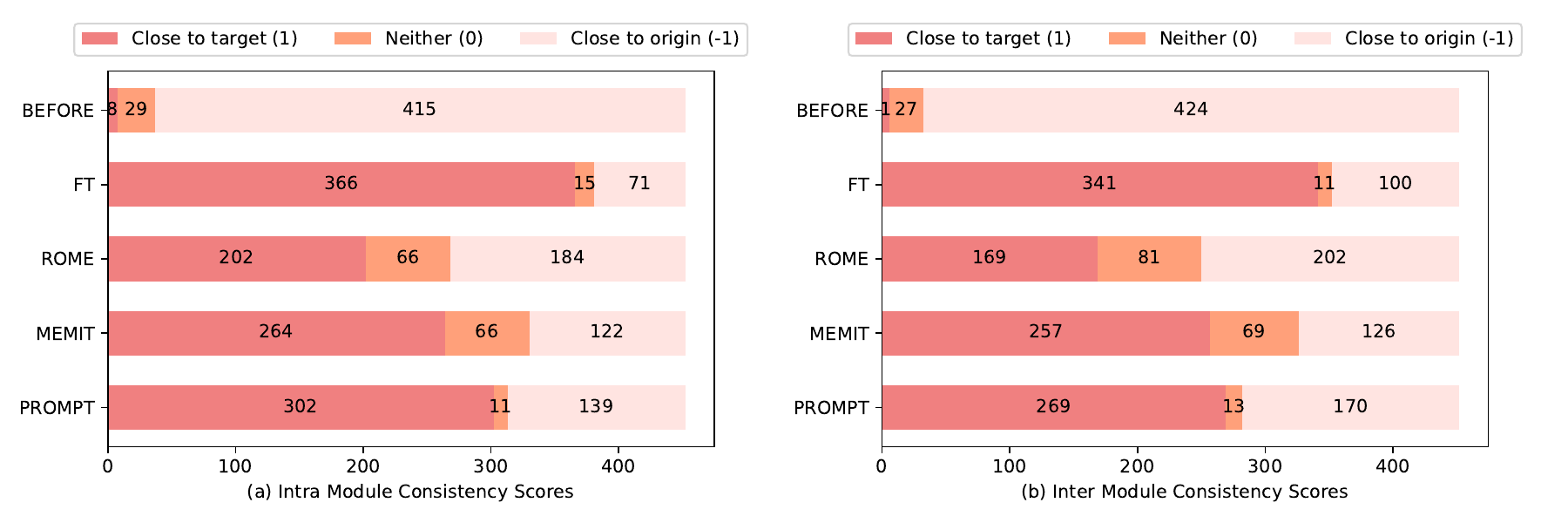}
    \caption{The results of the \metricone\ employed on the \llama\ across both intra and inter modules. 
    This investigation entailed a comparison of generated sentences both pre-edited and post-edited via different editing methods.
    The evidence clearly indicates that FT surpasses other methodologies.}
    \label{fig:congauge}
\end{figure*}

\subsection{Main Results}
The experimental results depicted in Table \ref{tab:maintable} provide a quantitative assessment of various editing methodologies on concept editing task.
1) Firstly, it is noteworthy that all methods tested in larger scale models demonstrate high \textbf{reliability}, indicating their potential utility in addressing modifications at the concept-level definition. 
FT shows notable reliability which achieves 100 percent, but limits to smaller model GPT2-XL. 
2) When shedding light on their adaptability to in-scope neighbors, there is still a discrepancy in performance; specifically, the results of \textbf{generalization} show a substantial decline when compared to reliability.
Moreover, larger scale models demonstrate enhanced generalization capabilities post-editing relative to their smaller counterparts.
Observations further reveal that method PROMPT stands out for its generalization and underscores its proficient understanding of conceptual context, even when prefix inputs are rephrased.
3) Additionally, the high \textbf{locality} results indicate that MEMIT exhibits the least impact on out-of-scope neighbors.
Such performance implies that MEMIT operates with greater precision in locating and modifying the necessary parameters.
4) ROME leads to the clearest variations on \textbf{instance change}, with a notable impact observed in LLaMA where about one-third of the instance-to-category relationships are modified, emphasizing the instance-level alterations due to conceptual knowledge change. 
Conversely, the application of PROMPT within GPT-J is unsuccessful on instance change, as discussed in Appendix \ref{appendix:special}.

\subsection{Analysis}

\paragraph{The gap between Reliability and \metricone\ signals the necessity for concept specific evaluation metrics.}
Figure \ref{fig:congauge} presents the outcomes of the \metricone, the novel metric established in \S \ref{sec:metric}.
We choose LLaMA for its high-quality text generation, which surpasses the other models in producing responses with fewer meaningless repetitions and incoherent statements.
% From the results, the majority of items initially receive a score of -1 before editing, which demonstrates the efficacy of the GPT-4 API in adhering to our predefined preferences.
Upon editing the conceptual knowledge, the FT outstrips other approaches on \metricone, with 366 items more closely with the intended definitions as opposed to 71 items retaining their original state. 
PROMPT method also results in the desired change of generated definition text in over half of the test samples.
\textbf{To demonstrate the alignment between GPT-4 evaluations and human preference within \metricone, we select 50 cases from the entire set for manual evaluation, which are included in Appendix \ref{Appendix:manual}.}

\begin{figure}
    \centering
    \includegraphics[width=0.99\linewidth]{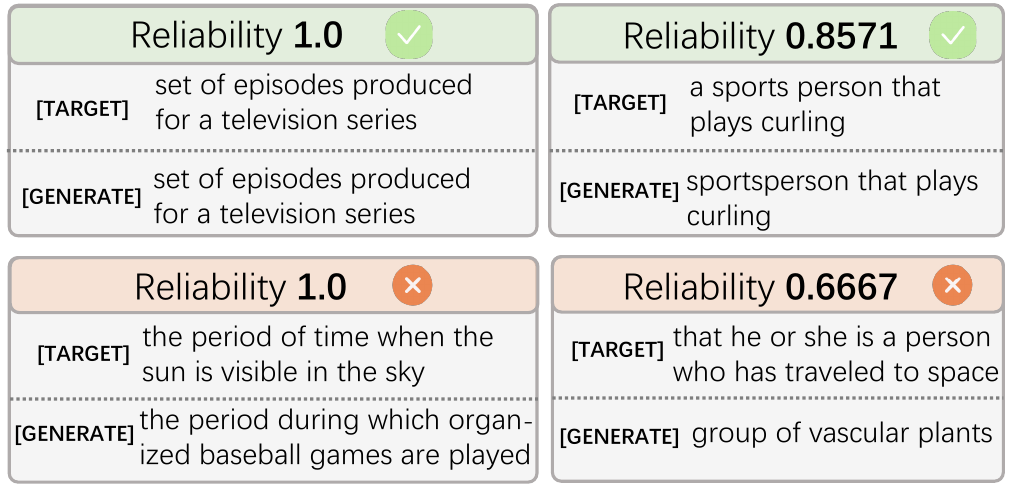}
    \caption{Cases of Reliability Scores vs Generated Sentences. 
    This Figure lists four representative cases that showcase the discrepancy.}
    \label{fig:case}
\end{figure}

\begin{figure*}
    \centering
    \includegraphics[width=0.96\linewidth]{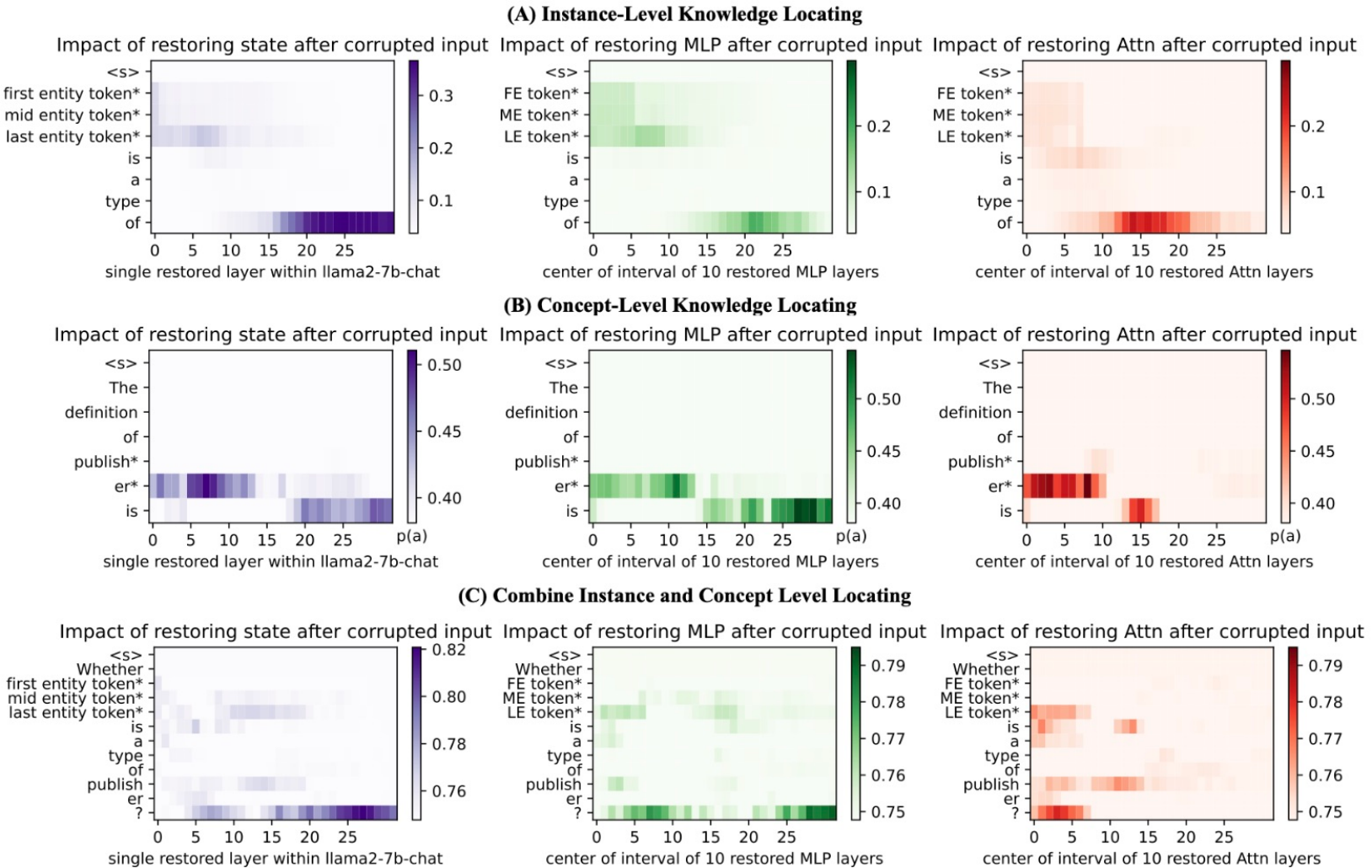}
    \caption{The conceptual and instantial knowledge locating in LLaMA-2-7B-Chat for the concept \textbf{publisher} and its corresponding instances by perturbing the input tokens.}
    \label{fig:locating_main}
\end{figure*}

\begin{figure}
    \centering
    \includegraphics[width=1.0\linewidth]{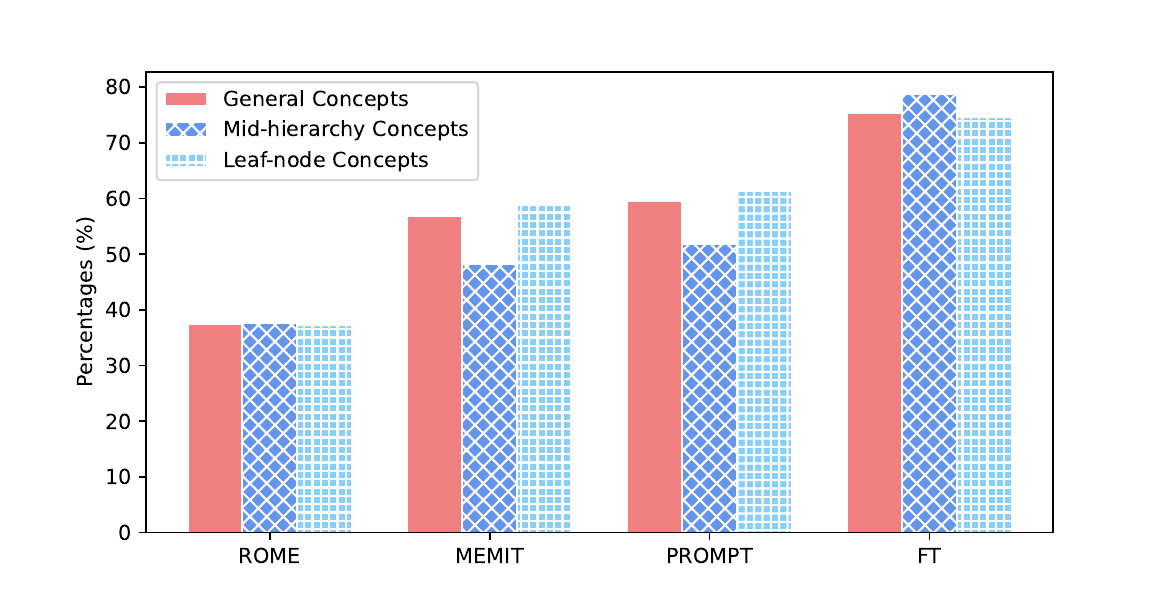}
    \caption{Considering concepts as tree-like structure, we assess the successful edits on mid-hierarchy and leaf-node concepts for more comprehensive analysis.
    The success rate is calculated by dividing the number of items get 1 in \metricone\ by the total items.}
    \label{fig:midnode}
\end{figure}

The generation function, which involves complex decoding mechanisms like probability normalization and sampling, creates an imperative for this evaluation, yielding text more diverse and coherent than selections from raw logit outputs.
Typically, in instance-level editing, the generation remains consistent with the target entities.
Although Reliability and the \metricone\ draw upon the same input, their performance in actual assessments diverges.
In Figure \ref{fig:case}, even if reliability achieves a perfect score of 1.0, where each token predicted exactly corresponds to the matching next token in the target text, this does not ensure that the generated text as a whole is an exact match to the target text. 
Conversely, a slightly lower reliability score does not imply a complete failure.
The reason is that each minor deviation during the generation process can cause the text to gradually diverge from the target.
This indicates that when editing long texts, it is necessary to account for the uncertainties of the entire text generation process, notwithstanding the precision of individual token predictions.

\paragraph{The impact of concepts' structure on editing effects across superclasses but NOT hierarchy.}

The comparison between intra and inter splits exposes another subtle yet important challenge in conceptual knowledge editing.  
Although results in Table \ref{tab:maintable} do not demonstrate significant differences in those metrics, Figure \ref{fig:congauge} illustrates a notable ease in \metricone\ when the definition is substituted with a concept from the same superclass, likely owing to the pre-existing higher-level connection of two concepts.
The findings indicate that mastering concepts spanning diverse superclasses tends to be moderately more challenging. 
Note that old metrics used to quantify the editing performance might not be sensitive enough to capture these disparities linked to superclass structures.
Meanwhile, even though some editing methods exhibit a slightly higher success rate on leaf-node concepts than mid-hierarchy ones in Figure~\ref{fig:midnode}, this minor gap does not substantially affect the overall effectiveness and there is no need for strategic adjustments based on hierarchical differences.

\paragraph{Generated sentence shows varying degrees of success in edits.}
For the concept editing task, the ultimate goal pursued is for the model-generated sentences to match the target exactly. 
In practice, we encounter a variety of situations that reflect the model's varying degrees of success and failure in executing editing instructions.
These categories and statements are detailed in Appendix \ref{appendix:cases}.

% Casual Tracing

\subsection{Locating Conceptual Knowledge in LLMs}
To further explore the storage patterns and mechanisms \cite{DBLP:journals/corr/abs-2402-10646, DBLP:journals/corr/abs-2403-00824} of correlation 
between concepts and instances, we follow~\citet{DBLP:conf/nips/MengBAB22}, identifying neurons that have
the strong causal effect in LLaMA-2-7B-Chat which has 32 transformer layers. 
The process of \textit{Causal Tracing} specifically involves three steps: \textit{clean run, corrupted run, corrupted-with-restoration run}.
It includes selecting certain specified tokens and recording the activation states before and after the addition of random noise, with the probability difference termed as the Indirect Effect (IE).
Detailed formulas are provided in Appendix~\ref{Appendix:knowledge_locating}.
We design three prompt variations:``[instance] is a type of'', ``The definition of [concept] is'' and ``Whether [instance] is a type of [concept]'' to probe the instantial and conceptual knowledge, and then perturb the instance and concept tokens respectively.
We carry out the analysis on the case concept ``publisher'' and average the hidden activation of all instances.

From Figure~\ref{fig:locating_main}(A), there is strong causality at a ‘late site’ in last few layers at the final token, in line with earlier studies about instances. 
Decomposing the effects into Multilayer Perceptron (MLP) and Attention (Attn) lookups, the observation for instantial knowledge reveals that MLP contributions are predominant at the ‘early site’, coupled with Attn at the ‘late site’ as the model retrieves its concept.
However, when locating conceptual knowledge as in Figure \ref{fig:locating_main}(B), it becomes apparent that both MLP and Attn module assume a heightened significance at ‘early site’.
At ‘late site’, the MLP shows greater importance in last few layers but Attn shows in middle layers. 
This could potentially explain the effectiveness of ‘locate-and-edit’ strategies when modify the definition but not be as adept in achieving instance change.
Results diverging from instantial knowledge may indicate the unique nature of concepts, with a high-dimensional generalization being more closely associated with attention in the early layers of the model.

Compared to the previous locating experiments, the association between conceptual knowledge and instantial knowledge may require the model to process in deeper exploration.
Therefore, we integrate both instance and concept tokens within a singular sequence that serves as the input for causal tracing.
From ~\ref{fig:locating_main}(C), although we can still observe that the last input token has the greatest influence on the entire response, the high IE performance in the attention layer has now shifted to the top ten layers.
This result suggests that attention mechanisms in earlier layers are more integral to the processing and representation of this instance-to-concept relationship.
To support the conclusions presented, we also carry out pertinent experiments on other cases, the details of which can be found in Appendix~\ref{Appendix:knowledge_locating}.

%% file: sections/related_work.tex
The current methods for knowledge editing are categorized into two main groups, those centered on preserving existing parameters and those entailing modification. 
The preservative methods incorporate explicit memory and prompting techniques to rectify model predictions. 
Examples include SERAC~\citep{DBLP:conf/icml/MitchellLBMF22}, MemPrompt~\citep{DBLP:conf/emnlp/MadaanTCY22} and IKE~\citep{DBLP:conf/emnlp/ZhengLDFWXC23}. 
Some modify the Feed-forward Neural Network (FFN) layer, as exemplified by CaliNET~\citep{DBLP:conf/emnlp/DongDSXSL22},T-Patcher~\citep{DBLP:conf/iclr/HuangSZZR023} and GRACE~\citep{hartvigsen2023aging}.
Alternatively, the locate-and-edit approaches need to first locate the relevant neurons, followed by the adjustment of corresponding target parameters. 
Representative studies are KN~\citep{DBLP:conf/acl/DaiDHSCW22}, ROME~\citep{DBLP:conf/nips/MengBAB22}, and MEMIT~\citep{meng2023massediting}.
Conversely, meta-learning utilize a hyper-network, a smaller network that generates the weights for layers in the main network, including KE~\citep{DBLP:conf/emnlp/CaoAT21}, MALMEN~\citep{DBLP:journals/corr/abs-2311-04661} and MEND~\citep{DBLP:conf/iclr/MitchellLBFM22}.

To facilitate the research of knowledge editing, numerous datasets are exploring the potentialities and far-reaching effects.
% MQUAKE \citep{DBLP:conf/emnlp/ZhongWMPC23} comprises multi-hop questions to test if edited models can correctly adjust answers based on changed facts.
% Later, RIPPLEEDITS \citep{DBLP:journals/corr/abs-2307-12976} extend their scope beyond single edited facts, delving into the verification on subsequent facts.
MQUAKE \citep{DBLP:conf/emnlp/ZhongWMPC23} challenges model updates to factual changes using multi-hop questions, 
while RIPPLEEDITS \citep{DBLP:journals/corr/abs-2307-12976}, DUnE \citep{DBLP:conf/emnlp/AkyurekPKW23} and ReCoE \citep{DBLP:journals/corr/abs-2401-17585} expand the scope to encompass reasoning over subsequent facts.
BAKE \citep{DBLP:journals/corr/abs-2310-10322} assesses the reversibility of editing.
Given the sequence of batch edits, 
\citet{li2023unveiling} identify paired edits that generate conflicts, and \citet{DBLP:conf/emnlp/0003ARC23} examine dependency within internal logical constraints.
Except for datasets which edit objects in triples, 
\citet{wei2023assessing} adopt a relation-centric perspective in edits.
\citet{hazra2024sowing} investigate how the edits impact the safety.
To the best of our knowledge, we are the first benchmark in LLMs for conceptual knowledge editing.

%% file: sections/tab/cases.tex
\begin{table*}[htbp]
\centering
\resizebox{2.05 \columnwidth}{!}{
\begin{tabular}{|c|m{1.1\linewidth}|}
\hline
\textbf{Category} & \multicolumn{1}{c|}{\textbf{Case in generated sentence}} \\
\hline
& group of people that play team handball. \\
\textbf{Case A} & \textbf{[TARGET]}: group of people that play team handball. \\
& \textbf{[ORIGIN]}: facility that makes wine.\\
\hline
& 100\% pure chemical compounds that are composed of two or more different elements. \\
\textbf{Case B} & \textbf{[TARGET]}: pure chemical substance consisting of two or more different chemical elements. \\
& \textbf{[ORIGIN]}: biomolecule consisting of chains of amino acid residues. \\
\hline
& the process of creating a detailed plan for the production of a radio or television program.\\
\textbf{Case C} & \textbf{[TARGET]}: organization responsible for production and transmission of radio and television programs.\\
& \textbf{[ORIGIN]}: business entity formed by one or more lawyers to engage in the practice of law. \\
\hline
& an individual who has the potential to participate in the sport of beach volleyball at a competitive level.\\
\textbf{Case D} & \textbf{[TARGET]}: prospective recipient of an award or position.\\
& \textbf{[ORIGIN]}: sportsperson who plays beach volleyball.\\
\hline
& a film that was released in 1945, directed by Michael Curtiz and starring Tom Neal, Ann Sheridan, and Edward G.\\
\textbf{Case E} & \textbf{[TARGET]}: minor planet of the inner Solar System; not a comet.\\
& \textbf{[ORIGIN]}: camp in which people are imprisoned or confined, commonly in large groups, without trial.\\
\hline
\end{tabular}}

\caption{Diverse scenarios showcasing the model's range of outcomes, from successful editing executions to cases of failure. \textbf{[TARGET]} denotes the revised description.  \textbf{[ORIGIN]} refers to the initial recognition prior to editing.}
\label{tab:cases}
\end{table*}